\def\year{2020}\relax
\documentclass[letterpaper]{article} 
\usepackage{aaai20}  
\usepackage{times}  
\usepackage{helvet} 
\usepackage{courier}  
\usepackage[hyphens]{url}  
\usepackage{graphicx} 
\usepackage{graphicx}  
\usepackage{amsmath}
\title{DCMN+: Dual Co-Matching Network for Multi-choice Reading Comprehension}
\author{Shuailiang Zhang,\textsuperscript{\rm 1,2,3} Hai Zhao,\textsuperscript{\rm 1,2,3}\thanks{$\ $Corresponding author. This paper was partially supported by National Key Research and Development Program of China (No. 2017YFB0304100), 	Key Projects of National Natural Science Foundation of China (U1836222 and 61733011).}$\ $ Yuwei Wu,\textsuperscript{\rm 1,2,3} Zhuosheng Zhang,\textsuperscript{\rm 1,2,3} 
	\\  \Large\textbf{Xi Zhou,\textsuperscript{\rm 4} Xiang Zhou\textsuperscript{\rm 4}}\\
	\textsuperscript{\rm 1}Department of Computer Science and Engineering, Shanghai Jiao Tong University \\
	\textsuperscript{\rm 2}Key Laboratory of Shanghai Education Commission for Intelligent Interaction \\ and Cognitive Engineering, Shanghai Jiao Tong University, Shanghai, China\\
	\textsuperscript{\rm 3}MoE Key Lab of Artificial Intelligence, AI Institute, Shanghai Jiao Tong University, Shanghai, China\\
	\textsuperscript{\rm 4}CloudWalk Technology, Shanghai, China\\
	{ zsl123@sjtu.edu.cn, zhaohai@cs.sjtu.edu.cn,}\\
	{ \{will8821,zhangzs\}@sjtu.edu.cn, \{zhouxi,zhouxiang\}@cloudwalk.cn}\\
}
\begin{document}

\maketitle

\begin{abstract} 
	Multi-choice reading comprehension is a challenging task to select an answer from a set of candidate options when given passage and question. Previous approaches usually only calculate question-aware passage representation and ignore passage-aware question representation when modeling the relationship between passage and question, which cannot effectively capture the relationship between passage and question. In this work, we propose dual co-matching network (DCMN) which models the relationship among passage, question and answer options bidirectionally. Besides, inspired by how humans solve multi-choice questions, we integrate two reading strategies into our model: (i) passage sentence selection that finds the most salient supporting sentences to answer the question,  (ii) answer option interaction that encodes the comparison information between answer options. DCMN equipped with the two strategies (DCMN+) obtains state-of-the-art results on five multi-choice reading comprehension datasets from different domains: RACE, SemEval-2018 Task 11, ROCStories, COIN, MCTest. 
\end{abstract}

\section{Introduction}
Machine reading comprehension (MRC) is a fundamental and long-standing goal of natural language understanding which aims to teach machine to answer question automatically according to given passage \cite{Hermann15,Rajpurkar-D16,NguyenRSGTMD16,character}. In this paper, we focus on multi-choice MRC tasks such as RACE \cite{Lai-2017} which requests to choose the right option from a set of candidate answers according to given passage and question. Different from MRC datasets such as SQuAD \cite{Rajpurkar-D16} and NewsQA \cite{newsqa} where the expected answer is usually in the form of a short span from the given passage, answer in multi-choice MRC is non-extractive and may not appear in the original passage, which allows rich types of questions such as commonsense reasoning and passage summarization, as illustrated by the example in Table \ref{table01}. 

\begin{table}[t!] 
	\begin{center}
		\begin{tabular}{|p{7.6cm}|}
			\hline
			\textbf{Passage}: \textit{Runners in a relay race pass a stick in one direction. However, merchants passed silk, gold, fruit, and glass along the Silk Road in more than one direction. They earned their living by traveling the famous Silk Road. ... \textbf{The Silk Road was made up of many routes, not one smooth path.} They passed through what are now 18 countries. The routes crossed mountains and deserts and had many dangers of hot sun, deep snow and even battles...} \\ 
			\hline
			\textbf{Question}: \textit{The Silk Road became less important because \_ }.\\    
			\quad A. \textit{it was made up of different routes}\\  
			\quad B. \textit{silk trading became less popular}\\
			\quad C. \bf{\textit{sea travel provided easier routes}}  \\ 
			\quad D. \textit{people needed fewer foreign goods} \\
			\hline
		\end{tabular}
	\end{center}
	\caption{\label{table01} An example passage with related question and options from RACE dataset. The ground-truth answer and the evidence sentences in the passage are in \textbf{bold}.}
\end{table}

\begin{figure*}[t!]
	\centering
	\includegraphics[width=.85\textwidth ]{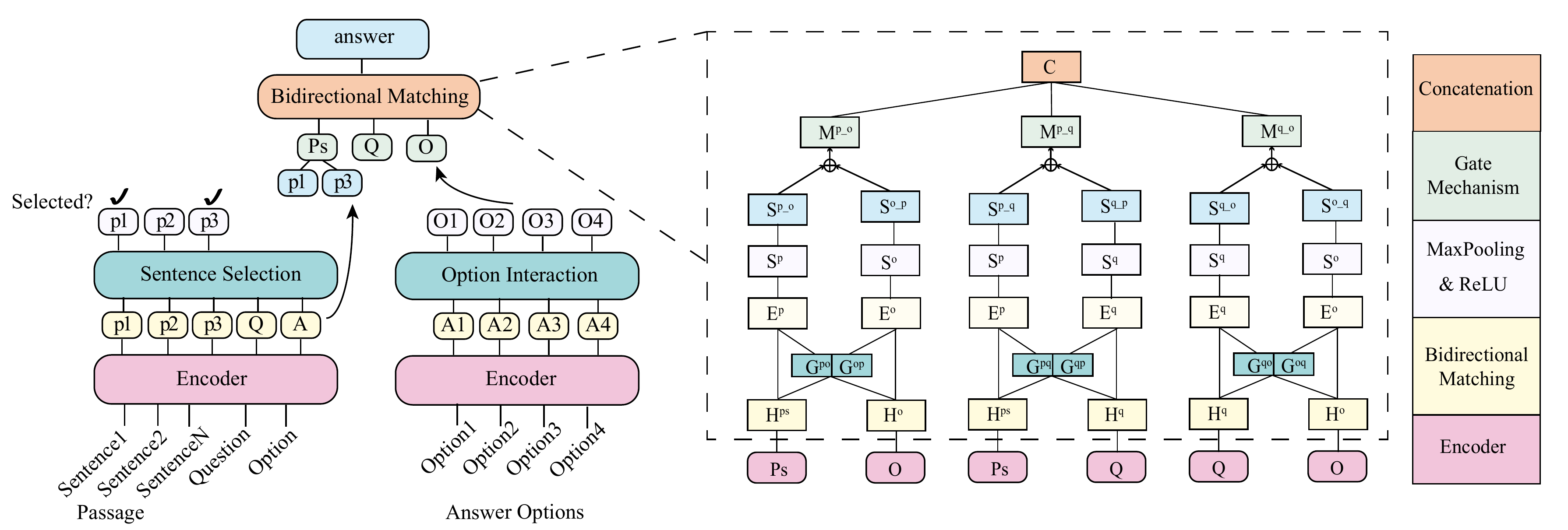}
	
	\caption{The framework of our model. P-Passage, Q-Question, O-Option.}
	\label{figure1}
\end{figure*}

Pre-trained language models such as BERT \cite{Devlin-18} and XLNet \cite{xlnet} have achieved significant improvement on various MRC tasks. Recent works on MRC may be put into two categories, training more powerful language models or exploring effective applying pattern of the language
models to solve specific task. There is no doubt that training a better language model is essential and indeed extremely helpful \cite{Devlin-18,xlnet} but at the same time it is time-consuming and resource-demanding to impart massive amounts of general knowledge from external corpora into a deep language model via pre-training \cite{SunKai-2018,sgnet}. For example, training a 24-layer transformer \cite{Devlin-18} requires 64 TPUs for 4 days. So from the practical viewpoint, given limited computing resources and a well pre-trained model, can we improve the machine reading comprehension during fine-tuning instead of via expensive full pre-training? This work starts from this viewpoint and focuses on exploring effective applying pattern of language models instead of presenting better language models to furthermore enhance state-of-the-art multi-choice MRC. We will show the way to use a strong pre-trained language model may still have a heavy impact on MRC performance no matter how strong the language model itself is. 

To well handle multi-choice MRC problem, an effective solution has to carefully model the relationship among the triplet of three sequences, passage (\textbf{P}), question (\textbf{Q}) and answer candidate options (\textbf{A}) with a matching module to determine the answer. However, previous unidirectional matching strategies usually calculate question-aware passage representation and ignore passage-aware question representation when modeling the relationship between passage and question \cite{Wang-2018,tang2019multi,Chen2018ConvolutionalSA}. 

To alleviate such an obvious defect in modeling the \{\textbf{P}, \textbf{Q}, \textbf{A}\} triplet from existing work, we propose dual co-matching network (DCMN) which bidirectionally incorporates all the pairwise relationships among the \{\textbf{P}, \textbf{Q}, \textbf{A}\} triplet. In detail, we model the passage-question, passage-option and question-option pairwise relationship simultaneously and bidirectionally for each triplet, exploiting the gated mechanism to fuse the representations from two directions. Besides, we integrate two reading strategies which humans usually use into the model. One is passage sentence selection that helps extract salient evidence sentences from the given passage, and then matches evidence sentences with answer options. The other is answer option interaction that encodes comparison information into each option. The overall framework is shown in Figure \ref{figure1}. The output of pre-trained language model (i.e. BERT \cite{Devlin-18} and XLNet \cite{xlnet}) is used as the contextualized encoding. After passage sentence selection and answer option interaction, bidirectional matching representations are built for every pairwise relationship among the \{\textbf{P}, \textbf{Q}, \textbf{A}\} triplet.

Our model achieves new state-of-the-art results on the multi-choice MRC benchmark challenge RACE \cite{Lai-2017}. We further conduct experiments on four representative multi-choice MRC datasets from different domains (i.e., ROCStories \cite{roc2016-corpus}, SemEval-2018 Task 11 \cite{semeval11}, MCTest \cite{mctest}, COIN Shared Task 1 \cite{semeval11}) and achieve the absolute improvement of 4.9\% and 2.8\% in average accuracy from directly fine-tuned BERT and XLNet, respectively, which indicates our method has a heavy impact on the MRC performance no matter how strong the pre-trained language model itself is.

\section{Our Proposed Model}
The illustration of our model is shown in Figure \ref{figure1}. The major components of the model are Contextualized Encoding, Passage Sentence Selection, Answer Option Interaction and Bidirectional Matching. We will discuss each component in detail.      

\subsection{Task Definition}

For the task of multi-choice reading comprehension, the machine is given a passage (\textbf{P}), a question (\textbf{Q}), and a set of answer candidate options (\textbf{A}) to select the correct answer from the candidates, where $\textbf{P} = \{\bf{p_1, p_2, ...,p_n}\}$ is the passage composed of $n$ sentences, $\textbf{A} = \{\bf{A_1, A_2, ...,A_m}\}$ is the option set with $m$ answer candidates.   

\begin{table*}[t!]
	
	\begin{center}
		\resizebox{.97\linewidth}{!}{
		\begin{tabular}{|p{0.8cm}|p{0.8cm}|p{0.8cm}|p{9cm}|p{4cm}|}
			\hline
			N sent & \% on RACE & \% on COIN &  Passage  & Question \\
			\hline
			1 & 65& 76& Soon after, the snack came out. \textit{I then opened the chips and started to enjoy them, before enjoying the \textbf{soda}.} I had a great little snack... & \textit{What else did the person enjoy?}    \\
			\hline
			2 & 22& 10& \textit{She lived in the house across the street that had 9 people and \textbf{3 dogs and another cat} living in it. She didn't seem very happy there, especially with a 2 year old that chased her and grabbed her.} The other people in the house agreed... & \textit{What did the 2 year old's mom own?}    \\
			\hline
			$>=$3 &13 & 14& \textit{When I was hungry last week for a little snack and a soda, I went to the closest vending machine. I felt that it was a little overpriced, but being as though I needed something...} & \textit{What's the main idea of this passage?}    \\
			\hline
		\end{tabular}}
	\end{center}
	\caption{\label{table2} Analysis of the sentences in passage required to answer questions on RACE and
		COIN. 50 examples from each dataset are sampled randomly. N sent indicates the number of sentences required to answer the question. The evidence sentences in the passage are in \textit{emphasis} and the correct answer is with \textbf{bold}.}
\end{table*}

\subsection{Contextualized Encoding} \label{Encoding}

In this work, pre-trained language models are used as the encoder of our model which encodes each token in passage and question into a fixed-length vector. Given an encoder, the passage, the question, and the answer options are encoded as follows:
\begin{equation} 
\begin{split} 
\textbf{H}^p=Encode(\textbf{P})&,\textbf{H}^q=Encode(\textbf{Q})\\ 
\textbf{H}^a=&Encode(\textbf{A})
\end{split}
\end{equation}
where $Encode(\cdot)$ returns the last layer output by the encoder, which can be well pre-trained language models such as BERT \cite{Devlin-18} and XLNet \cite{xlnet}, as using transformer as the contextualized encoder has shown to be very powerful in language representation \cite{sembert,zhou-zhao-2019-head,luo2019hierarchical,xiao}. $\textbf{H}^p \in R^{|P| \times l}$, $\textbf{H}^q \in R^{|Q| \times l}$, and $\textbf{H}^a \in R^{|A| \times l}$ are sequence representation of passage, question and answer option, respectively.
$|P|$, $|Q|$, $|A|$ are the sequence length, respectively. $l$ is the dimension of the hidden state.

\subsection{Passage Sentence Selection} \label{ss}
Existing multi-choice MRC models learn the passage representation with all the sentences in one-shot, which is inefficient and counter-intuitive. To explore how many sentences are necessarily required to answer the question, we randomly extract 50 examples from the development set of RACE and COIN, as shown in Table \ref{table2}. Among all examples, 87\% questions on RACE and 86\% on COIN can be answered within two evidence sentences. From this observation, the model should be extremely beneficial if focusing on a few key evidence sentences.

To select the evidence sentences from the passage $\textbf{P} = \{\bf{p_1}, p_2,..,p_i, ..,p_n\}$, this module scores each sentence $p_i$ with respect to the question $\textbf{Q}$ and answer option $\textbf{A}$ in parallel. The top $K$ scored sentences will be selected. This module shares the encoder with the whole model. For each \{$\bf{p_i}, \textbf{Q}, \textbf{A}$\} triplet, $\textbf{H}^{p_i} \in R^{|p_i| \times l}$, $\textbf{H}^q$, and $\textbf{H}^a$ are all representations offered by the encoder. Here we introduce two methods to compute the score of the triplet based on the representations.
\begin{itemize}
	\item \textbf{Cosine score}: The model computes word-by-word cosine similarity between the sentence and question-option sequence pair.
	\begin{equation} 
	\begin{split}
	\textbf{D}^{pa}&=Cosine(\textbf{H}^{a}, \textbf{H}^{p_i}) \in R^{|A| \times |p_i|} \\
	\textbf{D}^{pq}&=Cosine(\textbf{H}^{q}, \textbf{H}^{p_i}) \in R^{|Q| \times |p_i|} \\
	\bar{\textbf{D}}^{pa}&=MaxPooling(\textbf{D}^{pa}) \in R^{|A|} \\ 
	\bar{\textbf{D}}^{pq}&=MaxPooling(\textbf{D}^{pq}) \in R^{|Q|} \\
	score &= \frac{\sum_{k=1}^{|A|} \bar{\textbf{D}}^{pa}_k}{|A|} + \frac{\sum_{k=1}^{|Q|}  \bar{\textbf{D}}^{pq}_k}{|Q|}
	\end{split}
	\end{equation}
	where $\textbf{D}^{pa}$, $\textbf{D}^{pa}$ are the distance matrices and $\textbf{D}^{pa}_{ij}$ is the cosine similarity between the $i$-th word in the candidate option and the $j$-th word in the passage sentence.
	\item \textbf{Bilinear score}: Inspired by \cite{min-etal-2018-efficient}, we compute the bilinear weighted distance between two sequences, which can be calculated as follows:
	\begin{equation} 
	\begin{split}
	\alpha &=SoftMax(\textbf{H}^{q}W_1) \in R^{|Q| \times l} \\
	\textbf{q} &= \alpha^T\textbf{H}^{q} \in R^l \\
	\bar{\textbf{P}}_j &= \textbf{H}^{p_i}_jW_2\textbf{q} \in R^l, j \in [1,|p_i|] \\
	\hat{\textbf{P}}^{pq} &= Max(\bar{\textbf{P}}_1  \bar{\textbf{P}}_2,...,\bar{\textbf{P}}_{|p_i|}) \in R^l
	\end{split}
	\end{equation}
	where $W_1$, $W_2 \in R^{l \times l}$ are learnable parameters, $\hat{\textbf{P}}^{pq}$ is the bilinear similarity vector between the passage sentence and question. Similarly, the vector $\hat{\textbf{P}}^{pa}$ between the passage sentence and answer can be calculated with the same procedure. The final score can be computed as follows: 
	\begin{equation} 
	score = W_3^T\hat{\textbf{P}}^{pq} + W_4^T\hat{\textbf{P}}^{pa}
	\end{equation}
	where $W_3$, $W_4 \in R^l$ are learnable parameters. 
\end{itemize}

After scoring each sentence, top $K$ scored sentences are selected and concatenated together as an updated passage $\textbf{P}_s$ to replace original full passage. So the new sequence triplet is \{$\textbf{P}_s$, \textbf{Q}, \textbf{A}\} and the new passage is represented as $\textbf{H}^{ps}$.

\subsection{Answer Option Interaction} \label{oi}
Human solving multi-choice problem may seek help from comparing all answer options. For example, one option has to be picked up not because it is the most likely correct, but all the others are impossibly correct. Inspired by such human experience, we introduce the comparison information among answer options so that each option is not independent of the other. Here we build bilinear representations between any two options. Gated mechanism \cite{highway} is used to fuse interaction representation into the original answer option representations.

The encoder encodes each answer option $\textbf{A}_i$ as $\textbf{H}^{a_i}$. Then the comparison vector between option $\textbf{A}_i$ and $\textbf{A}_j$ can be computed as follows:  
\begin{equation} 
\begin{split}
\textbf{G} &=SoftMax(\textbf{H}^{a_i}W_5{\textbf{H}^{a_j}}^T) \in R^{|A_i| \times |A_j|}\\
\textbf{H}^{a_{i,j}} &=ReLU(\textbf{G}\textbf{H}^{a_j}) \in R^{|A_i| \times l}
\end{split}
\end{equation}
where $W_5 \in R^{l \times l}$ is one learnable parameter, $\textbf{G}$ is the bilinear interaction matrix between $A_i$ and $A_j$, $\textbf{H}^{a_{i,j}}$ is the interaction representation. Then gated mechanism is used to fuse interaction representation into the original answer option representations as follows:   
\begin{equation}
\begin{split}
\hat{\textbf{H}}^{a_{i}} &=[\{\textbf{H}^{a_{i,j}}\}_{j \ne i}] \in R^{|A_i| \times (m-1)l}\\
\bar{\textbf{H}}^{a_{i}} &=\hat{\textbf{H}}^{a_{i}}W_6 \in R^{|A_i| \times l} \\
g  &= \sigma(\bar{\textbf{H}}^{a_{i}}W_7 + \textbf{H}^{a_{i}}W_8 + b) \\
\textbf{H}^{o_i} &= g * \textbf{H}^{a_{i}}+(1-g) * \bar{\textbf{H}}^{a_{i}} 
\end{split} 
\end{equation}
where $W_7$, $W_8 \in R^{l \times l}$ and $W_6 \in R^{(m-1)l \times l}$ are learnable parameters, $\hat{\textbf{H}}^{a_{i}}$ is the concatenation of all the interaction representations. $g \in R^{|A_i| \times l}$ is a reset gate which balances the influence of $\bar{\textbf{H}}^{a_{i}}$ and $\textbf{H}^{a_{i}}$, and $\textbf{H}^{o_i}$ is the final option representation of $\textbf{A}_i$ encoded with the interaction information. At last, we denote $\textbf{O} = \{\textbf{H}^{o_1}, \textbf{H}^{o_2}, ...,\textbf{H}^{o_m}\}$ as the final answer option representation set fused with comparison information across answer options.

\subsection{Bidirectional Matching} \label{br} 
The triplet changes from \{\textbf{P}, \textbf{Q}, \textbf{A}\} to \{$\textbf{P}_s$, \textbf{Q}, \textbf{O}\} with passage sentence selection and answer option interaction. To fully model the relationship in the \{$\textbf{P}_s$, \textbf{Q}, \textbf{O}\} triplet, bidirectional matching is built to get all pairwise representations among the triplet, including passage-answer, passage-question and question-answer representation. Here shows how to model the relationship between question-answer sequence pair as an example and it is the same for the other two pairs.

Bidirectional matching representation between the question $\textbf{H}^q$ and answer option $\textbf{H}^o$ can be calculated as follows: 
\begin{equation} \label{eq1}
\begin{split}
\textbf{G}^{qo}&=SoftMax(\textbf{H}^qW_9{\textbf{H}^{o}}^T) \\
\textbf{G}^{oq}&=SoftMax(\textbf{H}^oW_{10}{\textbf{H}^{q}}^T)\\
\textbf{E}^{q}&=\textbf{G}^{qo}\textbf{H}^{o},
\textbf{E}^{o}={\textbf{G}^{oq}}\textbf{H}^{q}\\
\textbf{S}^{q} &= ReLU(\textbf{E}^{q}W_{11})\\
\textbf{S}^{o} &= ReLU(\textbf{E}^{o}W_{12})
\end{split}
\end{equation}
where $W_9$, $W_{10}$, $W_{11}$, $W_{12} \in R^{l \times l}$ are learnable parameters. $\textbf{G}^{qo} \in R^{|Q| \times |O|}$ and $\textbf{G}^{oq} \in R^{|O| \times |Q|}$ are the weight matrices between question and answer option. $\textbf{E}^{q} \in R^{|Q| \times l}, \textbf{E}^{o} \in R^{|A| \times l}$ represent option-aware question representation and question-aware option representation, respectively. The final representation of question-answer pair is calculated as follows:
\begin{equation} \label{eq2}
\begin{split}
\textbf{S}^{q\_o} &= MaxPooling(\textbf{S}^{q})\\
\textbf{S}^{o\_q} &= MaxPooling(\textbf{S}^{o})\\
g  &= \sigma(\textbf{S}^{q\_o}W_{13} + \textbf{S}^{o\_q}W_{14} + b) \\
\textbf{M}^{q\_o} &= g * \textbf{S}^{o\_q}+(1-g) * \textbf{S}^{o\_q}
\end{split}
\end{equation} 
where $W_{13}, W_{14} \in R^{l \times l}$ and $b \in R^{l}$ are three learnable parameters.  After a row-wise max pooling operation, we get the aggregation representation $\textbf{M}^{q} \in R^{l}$ and $\textbf{M}^{o} \in R^{l}$. $g \in R^l$ is a reset gate. $\textbf{M}^{q\_o} \in R^{l}$ is the final bidirectional matching representation of the question-answer sequence pair. 

Passage-question and passage-option sequence matching representation $\textbf{M}^{p\_q}, \textbf{M}^{p\_o} \in R^{l}$ can be calculated in the same procedure from Eq.(\ref{eq1}) to Eq.(\ref{eq2}). The framework of this module is shown in Figure \ref{figure1}.

\subsection{Objective Function}
With the built matching representations $\textbf{M}^{p\_q}, \textbf{M}^{p\_o}, \textbf{M}^{q\_o}$ for three sequence pairs, we concatenate them as the final representation $\textbf{C} \in R^{3l}$ for each passage-question-option triplet. We denote the representation $\textbf{C}_i$ for each \{$P_s, Q, O_i$\} triplet. If $A_k$ is the correct option, then the objective function can be computed as follows:
\begin{equation} 
\begin{split}
\textbf{C} &= [\textbf{M}^{p\_q}; \textbf{M}^{p\_o}; \textbf{M}^{q\_o}] \\
L(A_k|P,Q) &= -log{\frac{\text{exp}(V^T\textbf{C}_k)}{\sum_{j=1}^m{\text{exp}(V^T\textbf{C}_j)}}} 
\end{split}
\end{equation}
where $V \in R^{3l}$ is a learnable parameter and $m$ is the number of answer options. 

\section{Experiments}
\subsection{Dataset}
We evaluate our model on five multi-choice MRC datasets from different domains. Statistics of these datasets are detailed in Table \ref{table3}. Accuracy is calculated as $acc = N^+/N$, where $N^+$ and $N$ are the number of correct predictions and the total number of questions. Some details about these datasets are shown as follows:

\begin{table}[b]
	\begin{center}
		\begin{tabular}{p{1.5cm}p{2.7cm}p{0.3cm}p{0.8cm}p{0.8cm}}
			\hline
			Task& Domain &\#o &\#p &\#q \\
			\hline
			RACE &general &4 &27,933 &\textbf{97,687}\\
			SemEval &narrative text &2 &2,119 &13,939\\
			ROCStories &stories &2 &3472 &3472\\
			MCTest &stories &4 &660 &2,640\\
			COIN &everyday scenarios &2 &\_ &5,102\\
			\hline
		\end{tabular} 
	\end{center}
	\caption{\label{table3} Statistics of multi-choice machine reading comprehension datasets. \#o is the average number of candidate options for each question. \#p and \#q are the number of documents and questions in the dataset.}
\end{table}

\begin{itemize}
	\item \textbf{RACE} \cite{Lai-2017}: RACE consists of two subsets: RACE-M and RACE-H respectively corresponding to middle school and high school difficulty levels, which is recognized as one of the largest and most difficult datasets in multi-choice reading comprehension. 
	\item \textbf{SemEval-2018 Task11} \cite{semeval11}: Multi-choice questions should be answered based on narrative texts about everyday activities.
	\item \textbf{ROCStories} \cite{roc2016-corpus}: This dataset contains 98,162 five-sentence coherent stories in the training dataset, 1,871 four-sentence story contexts along with a right ending and a wrong ending in the development and test datasets, respectively.
	\item \textbf{MCTest} \cite{mctest}: This task requires machines to answer questions about fictional stories, directly tackling the high-level goal of open-domain machine comprehension.
	\item \textbf{COIN Task 1} \cite{semeval11}: The data for the task is short narrations about everyday scenarios with multiple-choice questions.
\end{itemize}

\subsection{Implementation Details}
Our model is evaluated based on the pre-trained language model BERT \cite{Devlin-18} and XLNet \cite{xlnet} which both have small and large versions. The basic version BERT$_\text{base}$ has 12-layer transformer blocks, 768 hidden-size, and 12 self-attention heads, totally 110M parameters. The large version BERT$_\text{large}$ has 24-layer transformer blocks, 1024 hidden-size, and 16 self-attention heads, totally 340M parameters. Two versions of XLNet have the similar sizes as BERT.

In our experiments, the max input sequence length is set to 512. A dropout rate of 0.1 is applied to every BERT layer. We optimize the model using BertAdam \cite{Devlin-18} optimizer with a learning rate 2e-5. We train for 10 epochs with batch size 8 using eight 1080Ti GPUs when BERT$_\text{large}$ and XLNet$_\text{large}$ are used as the encoder. Batch size is set to 16 when using BERT$_\text{base}$ and XLNet$_\text{base}$ as the encoder\footnote{Our code is at	https://github.com/Qzsl123/dcmn.}.

\begin{table}[t!]
	\begin{center}
		\resizebox{\linewidth}{!}{
			\begin{tabular}{lll}
				\hline
				\bf Model         &RACE-M/H &RACE\\
				\hline 
				HAF \cite{zhuhaichao2018hierarchical} &45.0/46.4 &46.0\\
				MRU \cite{tay} &57.7/47.4 &50.4\\
				HCM \cite{Wang-2018} &55.8/48.2 &50.4\\
				MMN \cite{tang2019multi} &61.1/52.2 &54.7\\
				GPT \cite{gpt} &62.9/57.4 &59.0\\
				RSM \cite{SunKai-2018}&69.2/61.5 &63.8\\
				OCN \cite{ocn}  &76.7/69.6 &71.7 \\
				XLNet \cite{xlnet} &85.5/80.2 &81.8 \\
				\hline
				BERT$_\text{base}$$^*$    &71.1/62.3 &65.0 \\
				BERT$_\text{large}$$^*$  &76.6/70.1 &72.0 \\
				XLNet$_\text{large}$$^*$  &83.7/78.6 &80.1 \\
				\hline
				Our Models&&\\
				BERT$_\text{base}$$^*$ + DCMN &73.2/64.2 &67.0\\
				BERT$_\text{large}$$^*$ + DCMN & 79.2/72.1 &74.1\\
				BERT$_\text{large}$$^*$ + DCMN + P$_\text{SS}$ + A$_\text{OI}$ & 79.3/74.4 &\textbf{75.8}\\
				XLNet$_\text{large}$$^*$ + DCMN + P$_\text{SS}$ + A$_\text{OI}$ &86.5/81.3 &\textbf{82.8}\\  
				\hline
				Human Performance&& \\
				Turkers     &85.1/69.4 &73.3\\
				Ceiling    &95.4/94.2 &94.5\\
				\hline
			\end{tabular}}
	\end{center}
	\caption{\label{tablerace} Experiment results on RACE test set. All the results are from single models. P$_\text{SS}$: Passage Sentence Selection; A$_\text{OI}$: Answer Option Interaction. $^*$ indicates our implementation.}
\end{table}

\subsection{Evaluation and Ablation Study on RACE}
Table \ref{tablerace} reports the experimental results on RACE and its two subtasks: RACE-M and RACE-H. In the table, Turkers is the performance of Amazon Turkers on a randomly sampled subset of the RACE test set and Ceiling is the percentage of the unambiguous questions with a correct answer in a subset of the test set. Here we give the
results of directly fine-tuned BERT$_\text{base}$, BERT$_\text{large}$ and XLNet$_\text{large}$ on RACE and get the accuracy of 65.0\%, 72.0\% and 80.1\%, respectively. Because of the limited computing resources, the largest batch size can only be set to 8 in our experiments which leads to 1.7\% decrease (80.1\% vs. 81.8\%) on XLNet compared to the result reported in \cite{xlnet}\footnote{The implementation is very close to the result 80.3\% in \cite{xlnet} when using batch size 8 on RACE.}.

The comparison indicates that our proposed method obtains significant improvement over pre-trained language models (75.8\% vs. 72.0\% on BERT$_\text{large}$ and 82.8\% vs. 80.1\% on XLNet$_\text{large}$) and achieves the state-of-the-art result on RACE.

In Table \ref{tableabrace}, we focus on the contribution of main components (DCMN, passage sentence selection and answer option interaction) in our model. From the results, the bidirectional matching strategy (DCMN) gives the main contribution and achieves further improvement by integrating with the two reading strategies. Finally, we have the best performance by combining all components.

\begin{table}[!t] 
	\centering
	\resizebox{.998\columnwidth}{!}{
		\begin{tabular}{llll}
			\hline
			&BERT$_\text{base}$ &BERT$_\text{large}$ &XLNet$_\text{large}$\\
			\hline
			base encoder &64.6 &71.8 &80.1\\
			\quad + DCMN &66.0 (+1.4) &73.8 (+2.0)  &81.5 (+1.4)\\
			\quad + DCMN + P$\_{SS}$ &66.6 (+2.0) &74.6 (+2.8) &82.1 (+2.0) \\
			\quad + DCMN + P$\_{OI}$ &66.8 (+2.2) &74.4 (+2.6) &82.2 (+2.1) \\
			\quad + DCMN + ALL (DCMN+) &\bf{67.4 (+2.8)} &\bf{75.4 (+3.6)} &\bf{82.6 (+2.5)} \\
			\hline
	\end{tabular}}
	\caption{\label{tableabrace} Ablation study on RACE dev set. P$_\text{SS}$: Passage Sentence Selection. A$_\text{OI}$: Answer Option Interaction. DCMN+: DCMN + P$_\text{SS}$ + A$_\text{OI}$.}
\end{table}

\begin{table*}[t!]
	\begin{center}
		\begin{tabular}{lcccccc}
			\hline
			Task&Previous STOA& &BERT &DCMN\_BERT &XLNet &DCMN\_XLNet \\
			\hline
			SemEval Task 11 &\cite{SunKai-2018} &89.5 &90.5 &91.8 (+1.3) &92.0 &93.4 (+1.4)\\
			ROCStories &\cite{SCT} &91.8 &90.8 &92.4 (+1.6) &93.8 &95.8 (+2.0)\\
			MCTest-MC160 &\cite{SunKai-2018} &81.7 &73.8 &85.0 (+11.2) &80.6 &86.2 (+5.6) \\
			MCTest-MC500 &\cite{SunKai-2018} &82.0 &80.4 &86.5 (+6.1)&83.4 &86.6 (+3.2)\\
			COIN Task 1&\cite{Devlin-18} &84.2 &84.3 &88.8 (+4.5)&89.1 &91.1 (+2.0)\\
			\hline
			\textbf{Average} &&85.8 &84.0 &\textbf{88.9 (+4.9)} &87.8 &\textbf{90.6 (+2.8)}\\
			\hline
		\end{tabular}
	\end{center}
	\caption{\label{tableall} Results on the test set of SemEval Task 11, ROCStories, MCTest and the development set of COIN Task 1. The test set of COIN is not public. DCMN\_BERT: BERT + DCMN + P$_\text{SS}$ + A$_\text{OI}$. Previous SOTA: previous state-of-the-art model. All the results are from single models.}
\end{table*}

\subsection{Evaluation on Other Multi-choice Datasets}
The results on four other multi-choice MRC challenges are shown in Table \ref{tableall}. When adapting our method to the non-conventional MRC dataset ROCStories which requires to choose the correct ending to a four-sentence incomplete story from two answer options \cite{roc2016-corpus}, the question context is left empty as no explicit questions are provided. Passage sentence selection is not used in this dataset because there are only four sentences as the passage. Since the test set of COIN is not publicly available, we report the performance of the model on its development set.

\begin{table*}[t!]
	\begin{center}
		\resizebox{.90\linewidth}{!}{
		\begin{tabular}{llllcl}
			\hline
			Model & RACE &Model &RACE &Model &RACE\\
			\hline 
			BERT$_\text{base}$ &64.6&&&&\\
			\hline
			+ Unidirectional&&&&& \\
			$[S^{P\_O}; S^{P\_Q}; S^{O\_Q}]$    &65.0 &	$[S^{P\_Q}; S^{Q\_O}]$  &63.4 &$[S^{P\_Q}; S^{O\_Q}]$  &64.5 \\
			$[S^{P\_O}; S^{Q\_P}; S^{O\_Q}]$    &65.2   &$[S^{P\_O}; S^{Q\_O}]$&63.6 &$[S^{Q\_P}; S^{O\_Q}]$& 65.2\\
			$[S^{P\_Q}; S^{P\_O}]$ (HCM)&64.4 &$[S^{P\_O}; S^{O\_Q}]$& 64.2&$[S^{P\_Q}; S^{O\_P}]$  &64.7\\
			
			$[S^{P\_O}; S^{P\_Q}; S^{Q\_O}]$ (HAF) &64.2&$[S^{P\_O}; S^{Q\_P}; S^{Q\_O}]$    &64.4&$[S^{Q\_P}; S^{Q\_O}]$& 64.3\\
			$[S^{Q\_O}; S^{O\_Q}; S^{P\_Q}; S^{P\_O}]$ (MMN) &63.2&&&&\\
			\hline
			+ Bidirectional&&&&&\\ 
			$[M^{P\_Q}; M^{P\_O}]$    &66.4 &$[M^{P\_O}; M^{Q\_O}]$    &66.0&$[M^{P\_Q}; M^{Q\_O}]$    &65.5\\
			$[M^{P\_Q}; M^{P\_O}; M^{Q\_O}]$ (DCMN) &\textbf{67.1}&&&&\\
			\hline
		\end{tabular}}
	\end{center}
	\caption{\label{tableablation} Performance comparison with different combination methods on the RACE dev set. (HCM) \cite{Wang-2018}, (HAF) \cite{zhuhaichao2018hierarchical}, (MMN) \cite{tang2019multi} are previous methods. We use BERT$_\text{base}$ as our encoder here. $[;]$ indicates the concatenation operation. $S^{P\_O}$ and $M^{P\_O}$ are the unidirectional and bidirectional representation referred in Eq. \ref{eq2}.}
\end{table*}

As shown in Table \ref{tableall}, we achieve state-of-the-art (SOTA) results on all datasets and obtain 3.1\% absolute improvement in average accuracy over the previous average SOTA (88.9\% vs. 85.8\%) by using BERT as encoder and 4.8\% (90.6\% vs. 85.8\%) by using XLNet as encoder. To further investigate the contribution of our model, we also report the results of directly fine-tuned BERT/XLNet on the target datasets. From the comparison, we can see that our model obtains 4.9\% and 2.8\% absolute improvement in average accuracy over the baseline of directly fine-tuned BERT (88.9\% vs. 84.0\%) and XLNet (90.6\% vs. 87.8\%), respectively. These results indicate our proposed model has a heavy impact on the performance no matter how strong the adopted pre-trained language model itself is.

\subsection{Comparison with Unidirectional Methods}
Here we focus on whether the bidirectional matching method works better than previous unidirectional methods. In Table \ref{tableablation}, we enumerate all the combinations of unidirectional matching strategies\footnote{Here we omit the combinations with $S^{O\_P}$ because we find the combinations with $S^{P\_O}$ works better than $S^{O\_P}$.} which only use passage-aware question representation $S^{Q\_P}$ or question-aware passage representation $S^{P\_Q}$ when modeling the relationship between the passage and question. Specially, we roughly summarize the matching methods in previous work (i.e. HCM, HAF, MMN) using our model notations which meet their general ideas except some calculation details.

From the comparison, we observe that previous matching strategies (HCM 64.4\%, HAF 64.2\%, MMN 63.2\%) fail to give further performance improvement over the strong encoder (64.6\%). In contrast, all bidirectional combinations work better than the encoder. All three pairwise matching representations ($M^{P\_Q}$, $M^{P\_O}$, $M^{Q\_O}$) are necessary and by concatenating them together, we achieve the highest performance (67.1\%).

\begin{table}[t!]
	\begin{center}
		\begin{tabular}{lllllll}
			\hline
			Top K& 1 &2 &3 &4 &5 &6 \\
			\hline
			RACE-cos  &58.4 &60.1 &63.3 &65.8 &\textbf{66.5} &66\\
			RACE-bi   &59.5 &60.5 &63.4 &\textbf{66.8} &66.4 &66.2\\
			\hline 
			COIN-cos  &81.0 &82.0 &\textbf{83.5} &83.0 &82.5 &82.4\\
			COIN-bi   &81.7 &82.0 &82.6 &\textbf{82.8} &82.4 &82.2\\
			\hline
		\end{tabular}
	\end{center}
	\caption{\label{tablescore} Results on RACE and COIN dev set with cosine and bilinear score in P$_\text{SS}$. We use BERT$_\text{base}$ as encoder here.}
\end{table}

\subsection{Results with Different Settings in P$_\text{SS}$}
Table \ref{tablescore} shows the performance comparison with different scoring methods, and we observe that both methods have their advantages and disadvantages. Cosine score method works better on COIN dataset (83.5\% vs. 82.8\%) and bilinear score works better on RACE dataset (66.8\% vs. 66.5\%).

\begin{figure}[t!]
	\centering
	\includegraphics[width=.95\columnwidth]{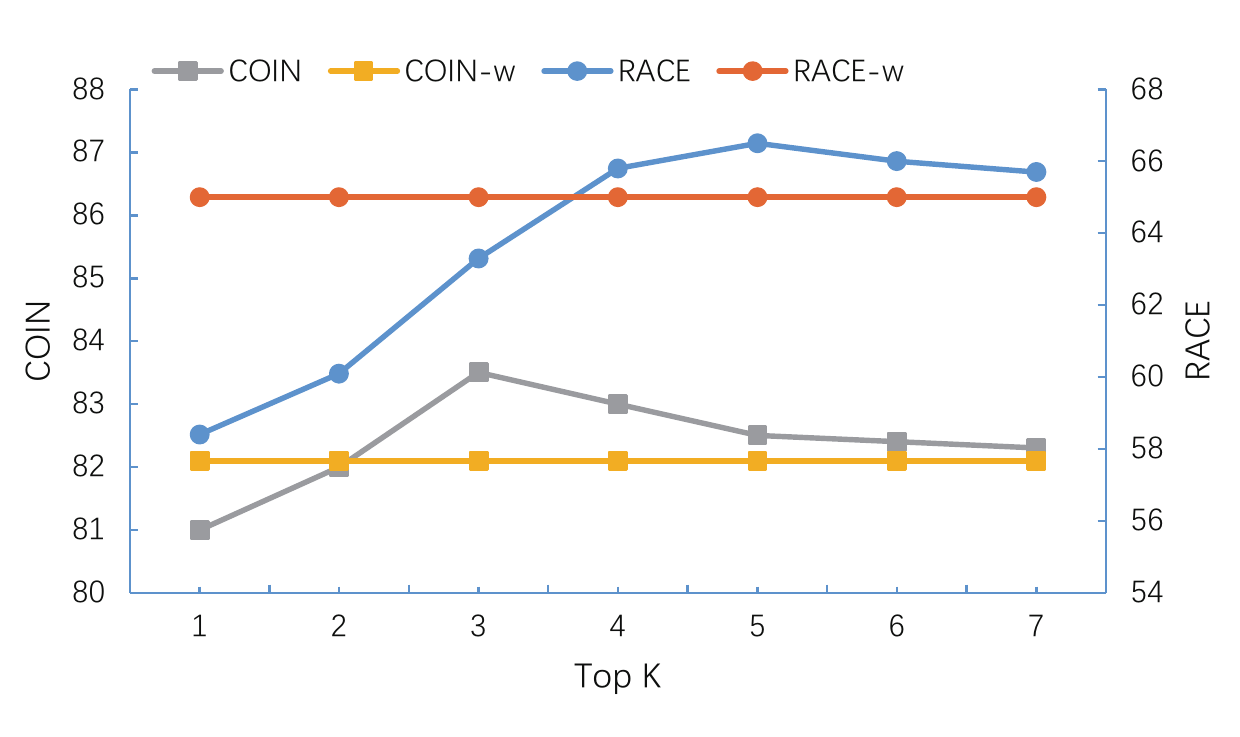}
	\caption{Results of sentence selection on dev sets of RACE and COIN when selecting different numbers of sentences (Top $K$). We use BERT$_\text{base}$ as encoder and cosine score method here. RACE/COIN-w indicates the results on RACE/COIN without passage sentence selection module.}
	\label{figuretop}
\end{figure}

Figure \ref{figuretop} shows the results of passage sentence selection (P$_{ss}$) on COIN and RACE dev set with different numbers of selected sentences (Top $K$). The results without P$_{ss}$ module are also shown in the figure (RACE-w and COIN-w). We observe that P$_{ss}$ mechanism consistently shows a positive impact on both datasets when more than four sentences are selected compared to the model without P$_{ss}$ (RACE-w and COIN-w). The highest performance is achieved when top 3 sentences are selected on COIN and top 5 sentences on RACE where the main reason is that the questions in RACE are designed by human experts and require more complex reasoning.  

\subsection{Why Previous Methods Break Down?} 
As shown in Table \ref{tableablation}, applying previous models to a strong BERT encoder fails to give performance increase over directly fine-tuned BERT. The contrast is clear that our proposed model achieves more than 3.8\% absolute increase over the BERT baseline. We summarize the reasons resulting in such contrast as follows: (i) the unidirectional representations cannot well capture the relationship between two sequences, (ii) previous methods use elementwise subtraction and multiplication to fuse $\textbf{E}^{q}$ and $\textbf{H}^{o}$ in Eq. \ref{eq1} (i.e., $[\textbf{E}^{q} \ominus \textbf{H}^{o}; \textbf{E}^{q} \otimes \textbf{H}^{o}]$) which is shown suboptimal as such processing breaks the symmetry of equation. Symmetric representations from both directions show essentially helpful for bidirectional architecture.

\subsection{Evaluation on Different Types of Questions}
\begin{figure}[ht]
	\centering
	\includegraphics[width=.95\columnwidth]{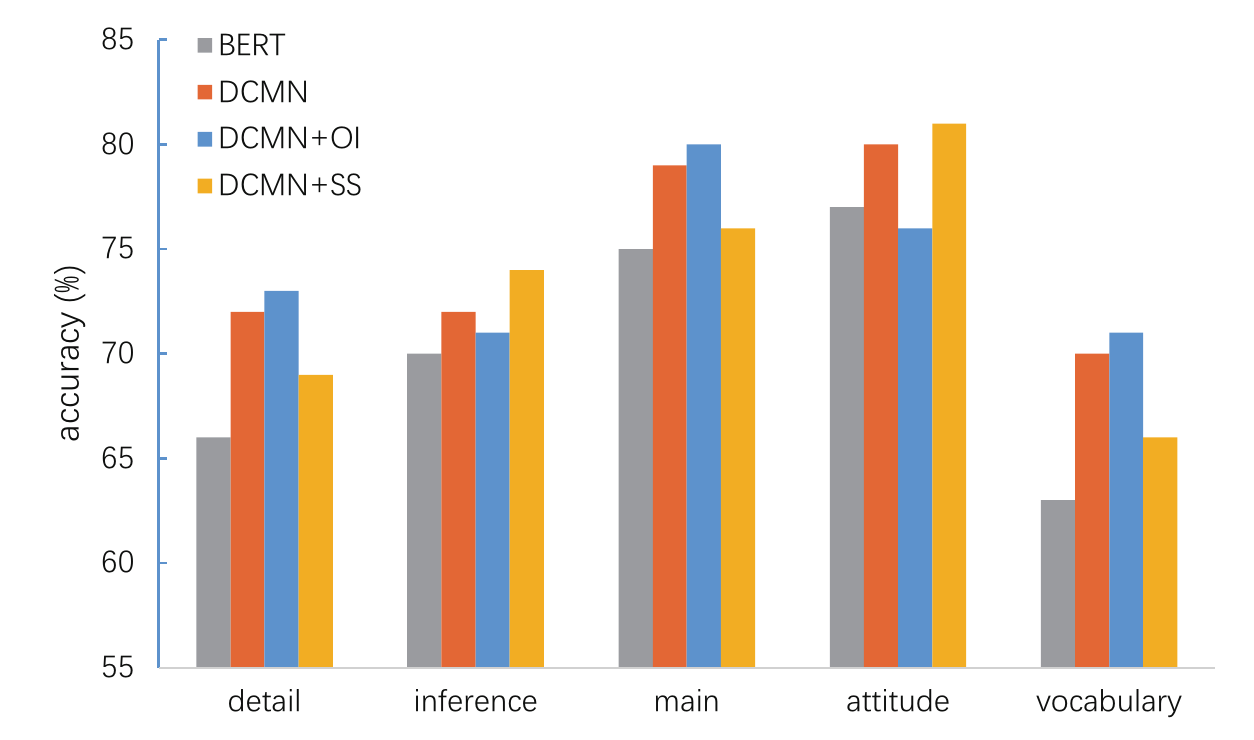}
	
	\caption{Results on different question types, tested on the RACE dev set. BERT$_\text{large}$ is used as encoder here. OI: Answer Option Interaction. SS: Passage Sentence Selection.}
	\label{figuretype}
\end{figure}

Inspired by \cite{SunKai-2018}, we further analyze the performance of the main components on different question types. Questions are roughly divided into five categories: \emph{detail}, \emph{inference}, \emph{main}, \emph{attitude} and \emph{vocabulary} \cite{Lai-2017,type}. We annotate all the instances of the RACE development set. As shown in Figure \ref{figuretype}, all the combinations of components work better than the BERT baseline in most question types. Bidirectional matching strategy (DCMN) consistently improves the results across all categories. DCMN+P$_\text{SS}$ works best on the \textit{inference} and \textit{attitude} categories which indicates P$_{SS}$ module may effectively improve the reasoning ability of the model. DCMN+A$_\text{OI}$ works better than DCMN on \textit{detail} and \textit{main} categories which indicates that the model achieves better distinguish ability with answer option interaction information.

\section{Related Work}
Neural network based methods have been applied to several natural language processing tasks, especially to MRC \cite{subseg,subword}.

The task of selecting sentences to answer the question has been studied across several question-answering (QA) datasets, by modeling the relevance between a sentence and the question 
\cite{Min-P18,evidence,choi,raiman-miller-2017-globally,yuanfudao}. \cite{evidence} apply distant supervision to generate imperfect labels and then use them to train a neural evidence extractor. \cite{Min-P18} propose a simple sentence selector to select the minimal set of sentences then feed into the QA model. They are different from our work in that (i) we select the sentences by modeling the relevance among sentence-question-option triplet, not sentence-question pair. (ii) Our model uses the output of language model as the sentence embedding and computes the relevance score using these sentence vectors directly, without the need of manually defined labels. (iii) We achieve a generally positive impact by selecting sentences while previous sentence selection methods usually bring performance decrease in most cases.

Most recent works attempting to integrate answer option interaction information focus on building attention mechanism at word-level \cite{ocn,zhuhaichao2018hierarchical,pujari} whose performance increase is very limited. Our answer option interaction module is different from previous works in that: (i) we encode the comparison information by modeling the bilinear representation among the options at sentence-level which is similar to modeling passage-question sequence relationship, other than attention mechanism. (ii) We use gated mechanism to fuse the comparison information into the original answer option representations.

\section{Conclusion}
This paper proposes dual co-matching network integrated with two reading strategies (passage sentence selection and answer option interaction) to enhance multi-choice machine reading comprehension. In terms of strong pre-trained language models such as BERT and XLNet as encoder, our proposed method achieves state-of-the-art results on five representative multi-choice MRC datasets including RACE. The experiment results consistently indicate the general effectiveness and applicability of our model.

\bibliography{AAAI20}
\bibliographystyle{aaai}

\end{document}